\documentclass[10pt,twocolumn,twoside]{IEEEtran}

\usepackage{stylefnl-bai}

\def\mgmm{$27.6\%~$}
\def\maddgmm{$30.5\%~$}

\title{\Large\bf Noisy Expectation-Maximization: Applications and Generalizations}

\author{Osonde Osoba, Bart Kosko\thanks{Prepared to the Black-in-AI workshop at NIPS, December 2017.

O. Osoba and B. Kosko are with the Signal and Image Processing Institute, Electrical Engineering Department, University of Southern California, Los Angeles, CA 90089-2564 USA (email: kosko@sipi.usc.edu)}}

\begin{document}
\maketitle

\begin{abstract}
We present a noise-injected version of the Expectation-Maximization (EM) algorithm: the Noisy Expectation Maximization (NEM) algorithm.  The NEM algorithm uses noise to speed up the convergence of the EM algorithm. The NEM theorem shows that injected noise speeds up the average convergence of the EM algorithm to a local maximum of the likelihood surface if a positivity condition holds. The generalized form of the noisy expectation-maximization (NEM) algorithm allow for arbitrary modes of noise injection including adding and multiplying noise to the data.

We demonstrate these noise benefits on EM algorithms for the Gaussian mixture model (GMM) with both additive and multiplicative NEM noise injection. A separate theorem (not presented here) shows that the noise benefit for independent identically distributed additive noise decreases with sample size in mixture models. This theorem implies that the noise benefit is most pronounced if the data is sparse. Injecting blind noise only slowed convergence.
\end{abstract}

\section{Noise Boosting the Expectation-Maximization Algorithm}

We show how carefully chosen and injected noise can speed convergence of the popular expectation-maximization (EM) algorithm.
A general theorem allows \emph{arbitrary} modes of combining signal and noise to improve the speed of parameter estimation.
The result still speeds EM convergence on average at each iteration so long as the injected noise satisfies a positivity condition.

The EM algorithm generalizes maximum-likelihood estimation to the case of missing or corrupted data\cite{dempster-laird-rubin1977,mclachlan-krishnan2007}.
Maximum likelihood maximizes the conditional signal probability density function (pdf) $f(y|\theta)$ for a random signal variable $Y$ given a vector of parameters $\theta$.
It equally maximizes the log-likelihood $\ln f(y | \theta)$ since the logarithm is monotone increasing.
So the maximum--likelihood estimate $\theta_*$ is
\begin{equation}
\theta_*= \argmax{\theta}\ \ln f(y | \theta).
\end{equation}
The parameter vector $\theta$ can contain means or covariances or mixture weights or any other terms that parametrize the pdf $f(y| \theta)$.
The data itself consists of observations or realizations $y$ of the signal random variable $Y$.
The data can be speech samples or image vectors or any type of numerical measurement.
The EM framework allows for missing or hidden data or so-called latent variables.
The random variable $Z$ denotes all such latent variables.
These latent variables can describe unseen states in a hidden Markov model or hidden neurons in a multilayer neural network.
Then $Z$ appears in the log-likelihood $\ln f(y | \theta)$ through the pdf identity $f(y|\theta) = \frac{f(y,z | \theta)}{f(z | y, \theta)}$.
This gives the key EM log-likelihood equality $\ln f(y | \theta)  = \ln f(y, z | \theta)- \ln f(z | y, \theta)$.

The EM algorithm estimates the missing information in $Z$  by iteratively maximizing the probability of $Z$ given both the observed data $y$ and the current parameter estimate $\theta_k$  \cite{Hogg2013}.
This involves averaging the log-likelihood $\ln f(y,z | \theta_k)$ over the conditional pdf $f(z|y, \theta_k)$ to form the surrogate likelihood function $Q(\theta |  \theta_k)$:
\begin{align}
Q(\theta|\theta_k) &= \E_{Z} \left[ \ln f(y,Z| \theta) \big{|} Y=y, \theta_k \right] \\
&= \int_{\mathcal{Z}} \ln [f(y,z| \theta)] f(z|y,\theta_k) ~\mathrm{d}z.
\end{align}
Then EM's ``ascent property" \cite{dempster-laird-rubin1977} uses Jensen's inequality \cite{cover-thomas91} and the above EM log-likelihood equality to ensure that any $\theta$ that increases the surrogate likelihood function $Q(\theta|\theta_k)$ can only increase the log--likelihood difference $\ln f(y | \theta)- \ln f(y | \theta_k)$:  $\ln\frac{f(y|\theta)}{f(y|\theta_k)} \ge Q(\theta | \theta_k)  -  Q(\theta_k | \theta_k)$.
The result is that EM is a hill-climbing algorithm that can only increase the log-likelihood at each step.

The EM algorithm iteratively climbs a hill of probability or log-likelihood until it reaches the closest peak of maximum likelihood.
The peak or mode corresponds to the locally maximal parameter $\theta_*$.
So the EM algorithm converges  to the local likelihood maximum $\theta_*$:  $\theta_k \rightarrow \theta_*$.
The EM algorithm halts in practice when its successive estimates $\theta_k$ differ by less than a given tolerance level $\|\theta_k-\theta_{k-1}\| < 10^{-tol}$ or when $|\ln f(y | \theta_k)- \ln f(y | \theta_{k-1})| < \varepsilon$ for some small positive $\varepsilon$.

The EM algorithm generalizes many popular algorithms.
These include the $k$-means clustering algorithm\cite{osoba-kosko2013} used in pattern recognition and big-data analysis, the backpropagation algorithm used to train deep feedforward and convolutional neural networks\cite{audhkhasi2013noise, audhkhasi2014noise, lecun2015}, and the Baum-Welch algorithm used to train hidden Markov models\cite{audhkhasi-osoba-kosko-HMM2013,welch2003}.
But the EM algorithm can converge slowly if the amount of missing data is high or if the number of estimated parameters is large\cite{mclachlan-krishnan2007,tanner1996}.
It can also get stuck at local probability maxima.
Users can run the EM algorithm from several starting points to mitigate the problem of convergence to local maxima.

The Noisy EM (NEM) algorithm~\cite{osoba-kosko2013,osoba-mitaim-kosko2013,osoba-mitaim-kosko2011,osoba-dissertation2013} is a noise-enhanced version of the EM algorithm that carefully selects noise and then \emph{injects} it into the data.
NEM converges faster on average than EM does because on average it takes larger steps up the same hill of probability or of log-likelihood.
NEM never takes shorter steps on average.
The largest noise gains tend to occur in the first few steps.
So NEM enhances the ascent property at each iteration.
This is a type of nonlinear noise benefit or \emph{stochastic resonance}\cite{wiesenfeld1995, bulsara1996, gammaitoni1998, mitaim-kosko1998SR, chapeau-blondeau-rousseau2004, lee2006, kosko2006, mcdonnell2008, patel-kosko-SPL2010, patel-kosko-TSP2011, chen2014, mitaim2014} that does not depend on a threshold \cite{franzke2011}.

NEM injects noise $N$ to the data $Y$  if the noise satisfies the NEM average positivity (nonnegativity) condition:
\begin{equation}
	\E_{Y,Z,N|\theta_*} \left[ \ln\left( \frac{f(\phi(Y,N),Z|\theta_k)}{f(Y,Z|\theta_k)} \right) \right] \geq 0\;. \label{eq:NEM-Goal}
\end{equation}
The NEM positivity condition~(\ref{eq:NEM-Goal}) holds when the noise-perturbed likelihood $f(\phi(y,N),z|\theta_k)$ is larger on average than the noiseless likelihood $f(y,z|\theta_k)$ at the $k^{th}$ step of the algorithm~\cite{osoba-mitaim-kosko2013,osoba-dissertation2013}.
This noise-benefit condition for additive noise injection has a simple quadratic form when the data or signal model is a mixture of Gaussian pdfs.

A simple argument gives the intuition behind the NEM positivity condition for additive noise.
This argument holds in much greater generality and underlies much of the theory of noise-boosting both the EM algorithm and Markov chain Monte Carlo algorithms \cite{Franzke2015}.
Consider a noise sample or realization $n$ that makes a signal $y$ more probable:  $f(\phi(y,n)|\theta) \geq f(y|\theta)$ for some parameter $\theta$.
The value $y$ is a realization of the signal random variable $Y$.
The value $n$ is a realization of the noise random variable $N$.
Then this pdf inequality holds if and only if $\ln\frac{f(\phi(y,n)|\theta)}{f(y|\theta)} \ge 0$.
Averaging over the signal and noise random variables gives the basic expectation form of the NEM positivity condition.
Averaging implies that the pdf inequality need hold only almost everywhere.
It need not hold on sets of zero probability.
This allows the user to ignore particular values when using continuous probability models.

Particular choices of the signal conditional probability $f(y|\theta)$ can greatly simplify the NEM sufficient condition.
This signal probability is the so-called ``data model" in the EM context of maximum likelihood estimation.
Estimation on Gaussian mixtures data models leads to simple quadratic NEM conditions. An exponential data model leads to an even simpler linear NEM condition.

%The same argument for multiplicative noise suggests that a simliar positivity condition should hold for a noise benefit. The mode of noise injection $\phi(y,n)$ includes noise addition $y+n$ and multiplication $y.n$ as long as the NEM noise condition is satisfied. There is nothing unique about the operations of addition or multiplication in this signal-noise context. So a noise benefit should hold for \emph{any} method $\phi(y, n)$ of combining signal $y$ and noise $n$ that obeys these pdf inequalities on average.

Theorem 1 presents the generalized form of the NEM Theorem for arbitrary measurable noise injection $\phi(Y,N)$. Corollaries 1 and 2 state the NEM sufficient condition for the special cases of additive and multiplicative injection: $\phi(Y,N) = Y+N$ and $\phi(Y,N) = YN$.

\begin{figure}[h]
\centerline{ \includegraphics[width=3.5in]{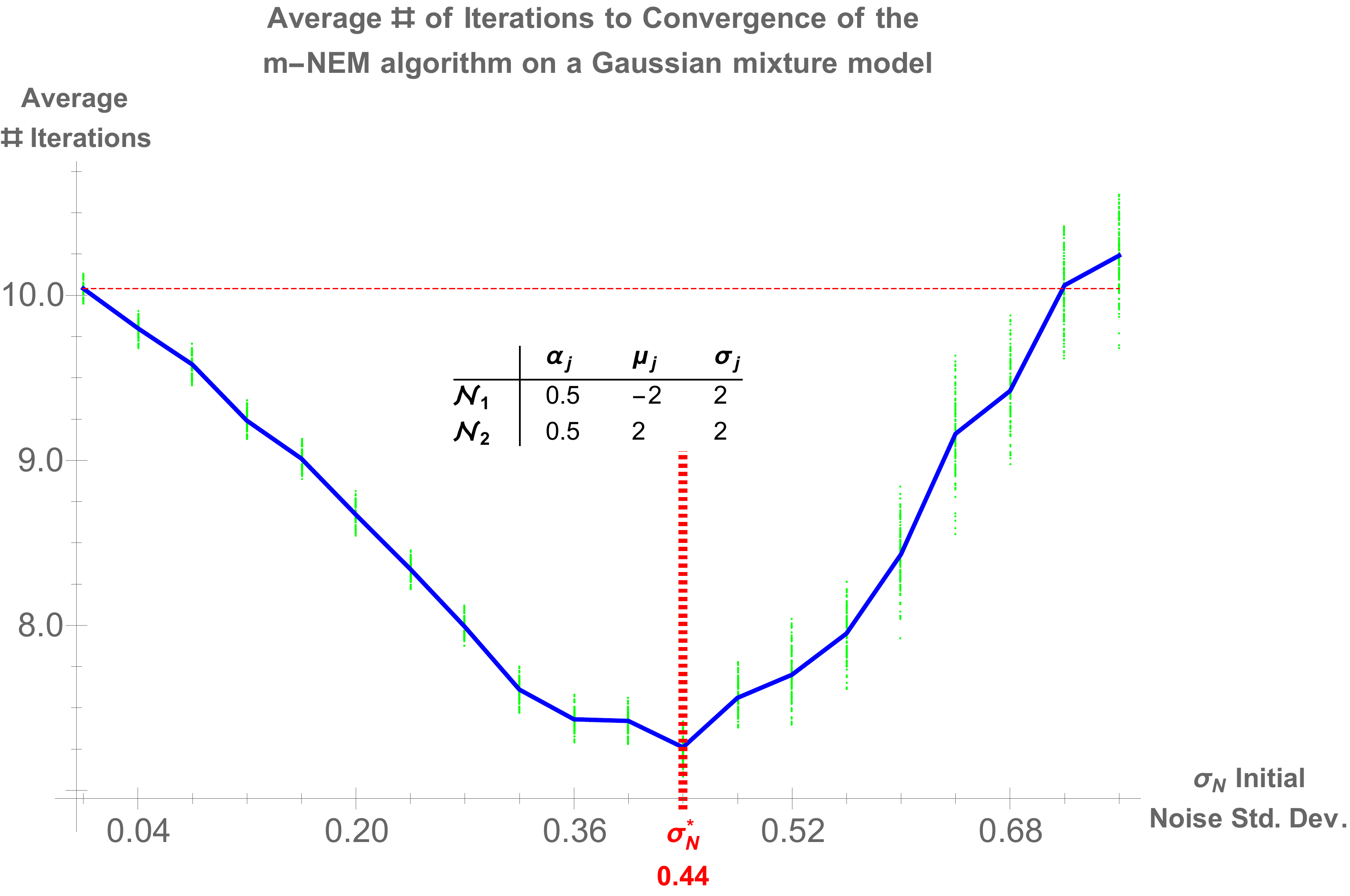} }
\caption{
{Multiplicative noise benefit when estimating the parameters of a sampled Gaussian mixture model.
The mixture density $f$ equally weighted two Gaussian probability density functions with the same variance of 4:  $f(x) =  \frac{1}{2}N_1(-2, 4)  +  \frac{1}{2}N_2(2, 4)$.
 The EM algorithm estimated the mixing weights, the means, and the variances of the two Gaussian densities.
Low intensity starting noise decreased the EM convergence time while higher intensity starting noise increased it.
The multiplicative noise had unit mean with different but decaying standard deviations.
The optimal initial noise standard deviation was $\sigma^* = 0.44$.
It gave a \mgmm speed-up over the noiseless EM algorithm.
Optimal m-NEM needed only 7 iterations on average to converge to the correct mixture parameters while noiseless EM needed 10 iterations on average.
The m-NEM procedure injected multiplicative noise that decayed at an inverse-square rate with the iterations.}
}
\label{fg:GaussNEM}
\end{figure}

Figure~\ref{fg:GaussNEM} shows an EM speed-up of \mgmm for multiplicative-NEM noise injection in the generic case of a bimodal mixture of two Gaussian pdfs.
Sampling from the mixture corresponds to sampling from two subpopulations that have the same variance but different means.
The task is threefold:  Estimate the unknown means of the two mixed Gaussian densities.
Estimate the unknown variances of the mixed densities.
And estimate the unknown mixture weights.
The mixture weights are nonnegative and sum to unity.

The noise-injected EM algorithm estimated all these parameters of the equally weighted two-pdf Gaussian mixture model.
Suppose random variable $X_j$ is Gaussian or normal with mean $\mu_j$ and variance $\sigma_j^2$:  $X_j  \sim N(\mu_j, \sigma_j^2)$ with pdf $f_j(x|\mu_j, \sigma_j^2)$.
Then the two-mixture density in Figure 1 had the form $f(x) = \alpha f_1(x|\mu_1, \sigma_1^2) + (1 - \alpha) f_2(x|\mu_2, \sigma_2^2)  =  \frac{1}{2} f_1(x| -2, 4)  +  \frac{1}{2} f_2(x|2, 4)$.
The data itself came from randomly samples of a Gaussian mixture.
The noise-boosted EM algorithm took on average only 7 iterations to estimate the Gaussian mixture parameters $\alpha, \mu_1, \mu_2, \sigma_1^2, \and \sigma_2^2$  while the noiseless EM algorithm took on average 10 steps.
The optimal initial noise standard deviation was $\sigma_N^* = 0.44$.
The simulations ``cooled" or ``annealed" the noise by multiplying the starting noise standard deviation $\sigma_N$ with the inverse-square term $k^{-2}$ at each iteration $k$.
This gradually shut off the noise injection as we discuss below when we present the details of the n-NEM algorithm.

Ordinary or \emph{blind}  noise (not subject to the appropriate NEM condition) only slowed EM convergence.
Blind noise was just noise drawn at random or uniformly from the set of all possible noise.
It was not subject to the NEM condition or to any other condition.

\begin{figure}[h]\centerline{ \includegraphics[width=3.5in]{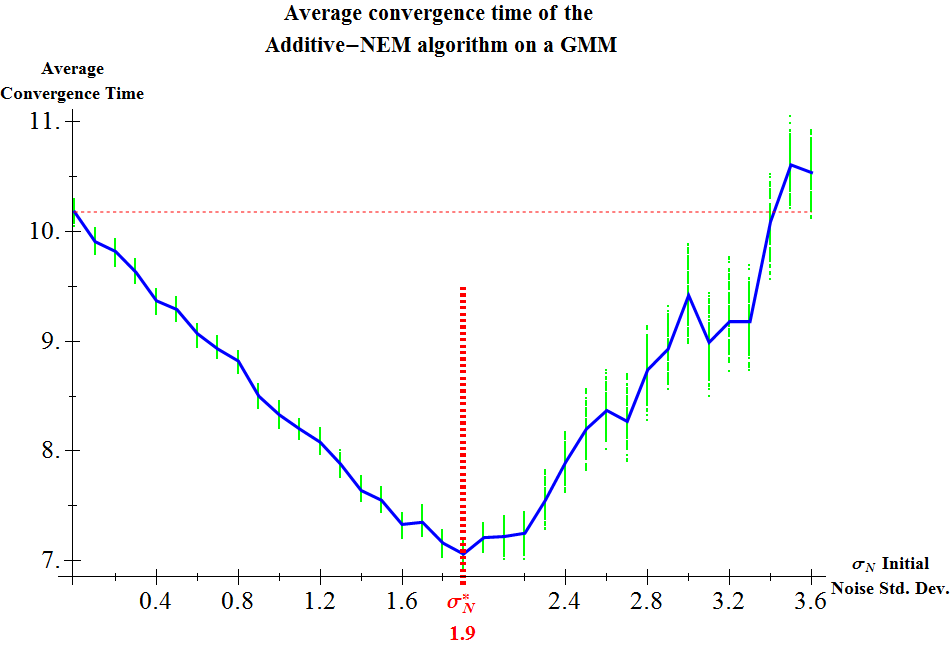} }\caption{{Noise benefit in the same GMM-NEM algorithm using additive noise injection. Low intensity noise decreases convergence time while higher intensity starting noise increases it. The noise decays at an inverse square rate. The optimal initial noise standard deviation is at $\sigma^* = 1.9$ which gives a \maddgmm speed improvement over the regular EM algorithm. This additive noise model results in slightly faster average convergence speed at the optimal noise level than the multiplicative noise model. But a $95\%$-bootstrap confidence interval for the average difference in optimal convergence time is $[-0.45,0.067]$. So the difference in optimal average convergence time is not statistically significant.}}\label{fg:GaussNEM-comp}\end{figure}

The optimal speed-up using additive noise on the same data model was \maddgmm at an optimal noise power of $\sigma^* = 1.9$. This speed-up was slightly better than the m-NEM speed-up for the same mixture model of two Gaussian pdfs. Figure~\ref{fg:GaussNEM-comp} shows the performance of the additive NEM algorithm on the same model.

A statistical test for the difference in the averaged optimal convergence times found that this difference was not statistically significant at the standard $0.05$ significance level.
Nor was it significant at the $0.10$ or $0.01$ levels.
The hypothesis test for the difference of means gave the very large bootstrap $p$-value (achieved significance level \cite{Hogg2013}) of $0.492$ based on 10,000 bootstraps.
That large $p$-value argues strongly against rejecting the null hypothesis that there was no statistically significant difference in the optimal average convergence times of the additive and multiplicative NEM speed-ups.

A $95\%$-bootstrap confidence interval for the average difference in optimal convergence time was $(-0.44, 0.06)$.
The confidence interval contained zero.
So we cannot reject the null hypothesis that the difference in optimal average convergence times for the two noise-injection modes was statistically insignificant at the $0.05$ level.
Nor can we reject the null hypothesis at the $0.10$ and $0.01$ significance levels because their respective $90\%$ and $99\%$ bootstrap confidence intervals were $(-0.40, 0.02)$ and $(-0.52, 0.13)$.
So there was no statistically significant difference in the noise speed-ups of the additive and mulitplicative cases.
An open and important research question is whether there are general conditions under which one of these noise injection modes outperforms the other.

\section{General Noise Injection for a NEM Benefit}
We next generalize the original proof for additive NEM~\cite{osoba-mitaim-kosko2013,osoba-dissertation2013} to NEM that uses an arbitrary mode of noise injection.
The metrical idea behind the proof remains the same: a noise benefit occurs on average at an iteration if the noisy pdf is closer to the optimal pdf than the noiseless pdf is.

Relative entropy measures the pseudo-distance of a pdf to the optimal pdf in a topological space of pdfs:

\begin{align}
	D\left(f(y,z|\theta_*)\Vert f_N(y,z|\theta_k)\right) \leq D\left(f(y,z|\theta_*)\Vert f(y,z|\theta_k)\right)
\end{align}
where
\begin{equation}
	f_N(y,z|\theta_k) = f(\phi(y, N),z|\theta_k)
\end{equation}
is the noise-injected pdf.
The literature sometimes refers to the relative entropy as the Kullback-Leibler divergence\cite{cover-thomas91}.

 The relative entropy is asymmetric and has the form of an average logarithm
\begin{equation}
D\left(h(u,v)||g(u,v)\right) = \int_{\mathcal{U}} \int_{\mathcal{V}} \ln \left[ \frac{h(u,v)}{g(u,v)} \right] h(u,v) ~\mathrm{d}u~\mathrm{d}v
\end{equation}
for positive pdfs $h$ and $g$ over the same support \cite{cover-thomas91}.
Convergent sums can replace the integrals in the discrete case.
We follow convention in calling the relative entropy a pseudo-metric.
It is technically only a pre-metric because the relative entropy between two pdfs is always nonnegative.
The relative entropy is zero if and only if the two pdfs are equal almost everywhere.
This yields the proof strategy of reducing the relative entropy with respect to the optimal pdf at each iteration $k$.

The key point is that the noise-injection mode $\phi(y, N)$ need be neither addition $\phi(y, N)=y+N$ nor multiplication $\phi(y, N)=yN$.
 It can be any measurable function $\phi$ of the data $y$ and the noise $N$.
This generality does not affect the main proofs for a noise benefit.

The above relative entropy inequality is logically equivalent to the EM noise-benefit condition at iteration $k$ if we cast the noise benefit in terms of expectations~\cite{osoba-mitaim-kosko2013}:

\begin{align}
\E \Big[ Q(\theta_*|\theta_*) - Q(\theta_{k}|\theta_*) \Big] &\geq \E \Big[ Q(\theta_*|\theta_*) - Q_N(\theta_{k}|\theta_*) \Big]  \label{eq:EQ-NoiseBenefit}
\end{align}
where $Q_N$ is the noise-perturbed surrogate likelihood function
\begin{equation}
	Q_N\left( \theta |\theta_k \right) = \E_{Z|Y,\theta_k}   \left[ \ln f_N(y,Z| \theta)  \right].
\end{equation}
Any noise $N$ that satisfies this EM noise-benefit condition will on average give better parameter estimates at each iteration than will noiseless estimates or those that use blind noise.
The relative-entropy version of the noise-benefit condition allows the same derivation of the generalized NEM condition as in the original case of additive noise.
The result is Theorem 1.

\noindent{\bf{Theorem 1: The Arbitrary-Injection NEM Theorem}}\label{thm:genNEM}\\ %\begin{thm}
 Let $\phi(Y,N)$ be an arbitrary mode of combining the signal $Y$ with the noise $N$.
Suppose the average positivity condition holds:
\begin{equation}
	\E_{Y,Z,N|\theta_*} \left[ \ln\left( \frac{f(\phi(Y, N),Z|\theta_k)}{f(Y,Z|\theta_k)} \right) \right] \geq 0\;.
	\label{eq:genNEM-Goal}
\end{equation}
Then the EM noise benefit
\begin{equation}
	\ Q(\theta_{k}|\theta_*)   \leq \  Q_N(\theta_{k}|\theta_*)
\end{equation}
holds on average at each iteration $k$:
\begin{multline}
\E_{Y| \theta_k} \Big[ Q \left( \theta_* |\theta_* \right)  -  Q \left( \theta_k |\theta_* \right) \Big] \geq \\ \E_{N,Y| \theta_k} \Big[ Q \left( \theta_* |\theta_* \right)  -  Q_N \left( \theta_k |\theta_* \right) \Big] \;.
\end{multline}
%\end{thm}

\noindent{\bf{Corollary 1: Additive NEM}}\label{thm:aNEM}\\
Suppose the average positivity condition holds for additive noise injection:
\begin{equation}
	\E_{Y,Z,N|\theta_*} \left[ \ln\left( \frac{f(Y+N,Z|\theta_k)}{f(Y,Z|\theta_k)} \right) \right] \geq 0\;.
	\label{eq:genNEM-Goal}
\end{equation}
Then the EM noise benefit
\begin{equation}
	\ Q(\theta_{k}|\theta_*)   \leq \  Q_N(\theta_{k}|\theta_*)
\end{equation}
holds on average at each iteration $k$:
\begin{multline}
\E_{Y| \theta_k} \Big[ Q \left( \theta_* |\theta_* \right)  -  Q \left( \theta_k |\theta_* \right) \Big] \geq \\ \E_{N,Y| \theta_k} \Big[ Q \left( \theta_* |\theta_* \right)  -  Q_N \left( \theta_k |\theta_* \right) \Big] \;.
\end{multline}

\noindent{\bf{Corollary 2: Multiplicative NEM (m-NEM)}}\label{thm:mNEM}\\
Suppose the average positivity condition holds for multiplicative noise injection:
\begin{equation}
	\E_{Y,Z,N|\theta_*} \left[ \ln\left( \frac{f(YN,Z|\theta_k)}{f(Y,Z|\theta_k)} \right) \right] \geq 0\;.
	\label{eq:genNEM-Goal}
\end{equation}
Then the EM noise benefit
\begin{equation}
	\ Q(\theta_{k}|\theta_*)   \leq \  Q_N(\theta_{k}|\theta_*)
\end{equation}
holds on average at each iteration $k$:
\begin{multline}
\E_{Y| \theta_k} \Big[ Q \left( \theta_* |\theta_* \right)  -  Q \left( \theta_k |\theta_* \right) \Big] \geq \\ \E_{N,Y| \theta_k} \Big[ Q \left( \theta_* |\theta_* \right)  -  Q_N \left( \theta_k |\theta_* \right) \Big] \;.
\end{multline}

The NEM Theorem and its corollaries give a general method for noise-boosting the EM algorithm.
Theorem 1 implies that on average these NEM variants outperform the noiseless EM algorithm.% This includes affine combinations.

Algorithm~\ref{algo:m-NEM} gives the {generalized--NEM} algorithm schema.
\noindent The operation \textsc{gNEMNoiseSample($\mathbf{y}, ~k^{-\tau}  \sigma_N$) } generates noise samples that satisfy the NEM condition for the current data model.
The noise sampling pdf depends on the vector of random samples $\mathbf{y}$ in the data-generating model.

\begin{algorithm}[H]
\caption{$\hat{\theta}_{gNEM}$ = gen-NEM-Estimate($\mathbf{y}$)}
\label{algo:m-NEM}
\begin{algorithmic}[1]
\REQUIRE $\mathbf{y} = \left( y_1,\ldots, y_M \right)$ : vector of observed incomplete data
\ENSURE $\hat{\theta}_{gNEM}$ : gNEM estimate of parameter $\theta$
\WHILE{($\|\theta_k-\theta_{k-1}\| \geq 10^{-tol}$)}
\STATE $\mathbf{N_S}${\bf -Step:} $\mathbf{n} \leftarrow $ NEMNoiseSample($\mathbf{y}, ~k^{-\tau}  \sigma_N$)
\STATE $\mathbf{N_G}${\bf -Step:} $\mathbf{y}_\dagger \leftarrow \phi(\mathbf{y}, \mathbf{n})$
\STATE {\bf E-Step:} $Q \left( \theta |\theta_k \right) \leftarrow\E_{Z|y,\theta_k}  \left[ \ln f(\mathbf{y}_\dagger, \mathbf{Z}|\theta)  \right] $
\STATE {\bf M-Step:} $\theta_{k+1} \leftarrow \argmax{\theta} \left\{ Q\left( \theta |\theta_k \right) \right\}$
\STATE $k \leftarrow k+1$
\ENDWHILE
\STATE $\hat{\theta}_{gNEM} \leftarrow \theta_k$
\end{algorithmic}
\end{algorithm}

\noindent The E-Step takes a conditional expectation of a function of the noisy data samples $\mathbf{y}_\dagger$ given the noiseless data samples $\mathbf{y}$.

A deterministic decay factor $k^{-\tau}$ scaled the noise on the $k^{th}$ iteration.
It did this by replacing the fixed standard deviation $\sigma_N$ of the noise with the weighted standard deviation $k^{-\tau}\sigma_N$.
So the NEM noise had slightly smaller standard deviation with each successive iteration.
$\tau$ was the noise decay rate~\cite{osoba-mitaim-kosko2013}.
The decay factor drove the noise $N_k$ to zero as the iteration step $k$ increased.
This eventually shut off the noise injection.
We found that the value $\tau = 2$ worked best in the simulations and thus we used an inverse-square scaling $k^{-2}$.

The inverse-square decay factor reduced the NEM estimator's jitter around its final value.
This was important because the EM algorithm converges to fixed-points.
Excessive estimator jitter prolongs convergence time even when the jitter occurs near the final solution.
Our simulations used the inverse-square and thus polynomial decay factor instead of the logarithmic cooling schedules found in annealing applications \cite{kirkpatrick-gelatt-vecchi1983, cerny1985, geman-hwang1986, hajek1988, kosko-nnfs}.

\section{Noise-Boosting Parameter Estimation for Gaussian Mixture Models}

Corollaries 1 and 2 from \cite{osoba-mitaim-kosko2013} lead to NEM conditions for GMMs because the noise condition applies to each mixed normal pdf in the mixture.
We state and prove the NEM GMM results for additive and multiplicative noise injection.
The resulting quadratic NEM conditions depend only on the Gaussian means and not on their variances.

A finite mixture model~\cite{redner-walker1984,mclachlan-peel2004,hogg-tanis2006, Hogg2013} is a convex combination of a finite number of similar pdfs.
So we can view a mixture as a convex combination of a finite set of similar sub-populations.
The sub-population pdfs are similar in the sense that they all come from the same parametric family.
Mixture models apply to a wide range of statistical problems in pattern recognition and machine intelligence.
A Gaussian mixture consists of convex-weighted normal pdfs.
The EM algorithm estimates the mixture weights as well as the means and variances of each normal pdf.
The GMM is by far the most common mixture model in practice \cite{gershenfeld1999}.
The EM algorithm offers a standard way to estimate the parameters of a mixture model.

Let $Y$ be the observed mixed random variable.
Let $K$ be the number of sub-populations.
Let $Z \in \left\{1,\ldots,K\right\}$ be the hidden sub-population index random variable.
The convex population mixing proportions $\alpha_1,\ldots, \alpha_K$ define a discrete pdf for $Z$: $P(Z=j) = \alpha_j$.
The pdf $f(y|Z=j,\theta_j)$ is the pdf of the $j^{th}$  sub-population where $\theta_1,\ldots, \theta_K $ are the pdf parameters for each sub-population.
The sub-population parameter $\theta_j$ can represent the mean or variance of a normal pdf or both.
It can represent any number of quantities that parametrize the pdf.

Let $\Theta$ denote the vector of all model parameters:  $\Theta = \left\{\alpha_1, \ldots , \alpha_K, \theta_1, \ldots , \theta_K\right\}$.
The mixing weights $\alpha_1, \ldots , \alpha_K$ are convex coefficients.
So they are nonnegative and add to unity.
And thus they define a discrete probability distribution.
The joint pdf $f(y, z|\Theta)$ is
\begin{equation}
f(y,z|\Theta) =\sum_{j=1}^K \alpha_j ~f(y|j,\theta_j) ~\delta[z-j] \;
\end{equation}
where $\delta[z-j] = 1$ if $z = j$ and $\delta[z-j] = 0$ otherwise.
The $K$ pdfs $f(y|j,\theta_j)$ are the mixed pdfs in the finite mixture.
Their structure determines the sufficient condition for a NEM noise benefit.

EM algorithms for finite mixture models estimate $\Theta$ using the sub-population index $Z$ as the latent variable.
The GMM-EM algorithm uses the following $Q$-function
\begin{align}
Q(\Theta|\Theta_k) =& \E_{Z|y,\Theta_k}[\ln f(y,Z|\Theta)]  \\
%=& \sum_{z} \left(\sum_j \delta[z-j] \ln[\alpha_j f(y|j,\theta_j)]\right) p_Z(z|y,\Theta_k) \\
=& \sum_{j} \ln[\alpha_j f(y|j,\theta_j)] p_Z(j|y,\Theta_k).
\label{eq:Q_Mixture}
\end{align}
We can now state and prove a sufficient condition involving the mixed pdfs $f(y|j,\theta_j)$  for an m-NEM noise benefit in a Gaussian mixture model.
The condition has a simple quadratic form.

\vspace{8pt}

%%%%%%%%%%%%%
\noindent{\bf{Corollary: NEM Condition for Gaussian Mixture Models}}\label{cor:GMM-NEM} \\
Suppose $Y|_{Z=j} \sim {\cal N}(\mu_j,\sigma^2_j)$ and so $f(y|j,\theta)$ is a normal pdf. Then the pointwise pdf noise benefit for additive noise
\begin{align}
		f(y + n|\theta) \geq f(y|\theta) %\Delta f_j(y,n) &\geq 0
\end{align}
holds if and only if
\begin{align}
n^2 \leq 2n\left(\mu_j-y\right) \;.
\label{eq:GaussMLECondn}
\end{align}

\noindent{\bf{Corollary: m-NEM Condition for Gaussian Mixture Models}}\label{cor:mGMM-NEM} \\
	Suppose that $Y|_{Z=j} \sim {\cal N}(\mu_j,\sigma^2_j)$.
So $f(y|j,\theta)$ is a normal or Gaussian pdf.
Then the pointwise pdf noise benefit for multiplicative noise
	\begin{align}
		f(y n|\theta) \geq f(y|\theta) %\Delta f_j(y,n) &\geq 0
	\end{align}
	holds if and only if
	\begin{equation}
		y(n-1)\left[ y(n+1) - 2\mu_j\right] \leq 0 \;.
		\label{eq:mGMM-NEMCond}
	\end{equation}

\section{Conclusion}

This discussion summarizes some of the basic theorems for speeding up EM algorithms using noise injection. The theorems apply for generalized noise injection modes. We present the specializations to additive and multiplicative noise injection. Our subsequent work has demonstrated the many supervised and unsupervised machine learning algorithms are special cases of EM algorithms. This means they benefit in speed and accuracy from the principled injection of noise (i.e. noise that satisfies the NEM condition). We have demonstrated such noise benefits in unsupervised learning (like clustering\cite{osoba-kosko2013} and hidden markov model training\cite{audhkhasi-osoba-kosko-HMM2013,welch2003}) and in backpropagation training for neural networks\cite{audhkhasi2013noise, audhkhasi2014noise}.

Open research questions include the determination of optimal injective noise, conditions under which either multiplicative or additive noise outperforms the other, and the effect of data sparsity on m-NEM speed-ups and other general modes of NEM-based noise injection.

\bibliographystyle{IEEEtran}
%\bibliography{all}

\end{document}